\documentclass[journal]{IEEEtran}

\usepackage{times}
\usepackage{epsfig}
\usepackage{graphicx}
\usepackage{subfigure}
\usepackage{amsmath}
\usepackage{amssymb}
\usepackage{algorithm}
\usepackage{algpseudocode}
\usepackage{verbatim}
\usepackage{balance}
\usepackage{color}
\usepackage[noadjust]{cite}

\def\bi{\begin{itemize}}
\def\ei{\end{itemize}}
\def\bequ{\begin{equation}}
\def\eequ{\end{equation}}

\def\benum{\begin{enumerate}}
\def\eenum{\end{enumerate}}

\begin{document}

\title{YoTube: Searching Action Proposal via Recurrent and Static Regression Networks}

%	Institution1\\
%	Institution1 address\\
%	{\tt\small firstauthor@i1.org}
%	% For a paper whose authors are all at the same institution,
%	% omit the following lines up until the closing ``}''.
%	% Additional authors and addresses can be added with ``\and'',
%	% just like the second author.
%	% To save space, use either the email address or home page, not both
%	\and
%	Second Author\\
%	Institution2\\
%	First line of institution2 address\\
%	{\tt\small secondauthor@i2.org}
%}
\author{Hongyuan~Zhu$^\star$,
	Romain Vial$^\star$,
	Shijian Lu,\\
	Yonghong Tian,
	Xianbin Cao
	\thanks{H.~Zhu and S.~Lu are with Institute for Infocomm Research, A*Star, Singapore (email: \{zhuh, slu\}@i2r.a-star.edu.sg).}
	\thanks{R.~Vial is with the Mines ParisTech, France (e-mail:romain.vial@mines-paristech.fr).}
	\thanks{Y.~Tian is with National Engineering Laboratory for Video Technology (NELVT), School of EECS, Peking University, Beijing, China (email:yhtian@pku.edu.cn)}
	\thanks{X.~Cao is with School of Automation Science and Electrical Engineering, Beihang University, Beijing, China (email:xbcao@buaa.edu.cn)}
	\thanks{This work was mainly done when R.~Vial was interned with the Institute for Infocomm Research. H.~Zhu and R.~Vial are equally contributed (\it{Corresponding authors: H~Zhu and R.~Vial}).}
}

\maketitle
%\thispagestyle{empty}

%%------------------------
\begin{figure*} [htb]
	\begin{center}
		\includegraphics[width = 0.9\textwidth]{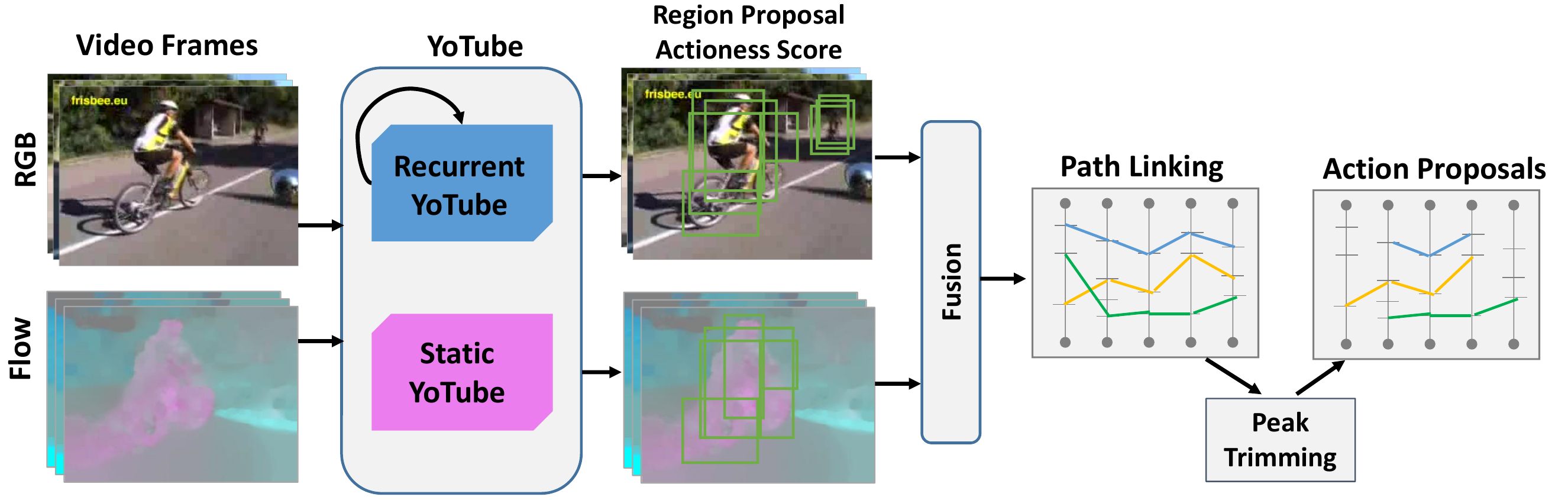}
	\end{center}
	\caption{\label{fig:flowmap}Conceptual illustration of our method: we explore the regression capability of RNN and CNN to directly regress sequences of bounding boxes. The located bounding boxes are then seamed into longer action proposals with path linking and trimming. }
\end{figure*}
%------------------------
%-------------------------------------------------------------------------

%%%%%%%%% ABSTRACT
\begin{abstract}
	In this paper, we present \textit{YoTube}-a novel network fusion 
	framework for searching action proposals in 
	untrimmed videos, where each action proposal corresponds to 
	a spatial-temporal video tube that potentially locates one human action. 
	Our method consists of a recurrent \textit{YoTube} detector and a static \textit{YoTube} detector, where
	the recurrent \textit{YoTube} explores the regression capability
	of RNN for candidate bounding boxes predictions using
	learnt temporal dynamics and the static \textit{YoTube}
	produces the bounding boxes using rich appearance cues
	in a single frame. Both networks are trained using 
	rgb and optical flow in order to fully exploit the rich  
	appearance, motion and temporal context, 
	and their outputs are fused to produce accurate 
	and robust proposal boxes. Action proposals are finally 
	constructed by linking these boxes using
	dynamic programming with a novel trimming method to 
	handle the untrimmed video effectively and efficiently. Extensive experiments on the challenging
	UCF-101 and UCF-Sports datasets show that our proposed technique 
	obtains superior performance compared with the state-of-the-art.
\end{abstract}

%%%%%%%%% BODY TEXT
\section{Introduction}
Video activity analysis has attracted increasing attention
in recent years due to its wide application in 
video surveillance and human computer interaction. 
Most works focus on action classification~\cite{LRCN_DonahueHGRVDS15,2Stream_WangSWVH16},
which aims to assign a global category label to a video sequence. 
On the other side, action detection~\cite{Jain_cvpr14,Oneata_eccv14,FindingTube_GkioxariM15}
which targets to localize the spatial-temporal extent of 
human actions is equally important by providing localized action information
for tasks requiring precise positioning e.g. assistive
agents and autonomous vehicles. 
Similar to the object detection where fast and reliable object proposal 
can greatly improve the performance of higher level understanding tasks, 
an efficient action proposal can greatly facilitate video activity 
analysis~\cite{YuYL13, FanSW13, ZhaoYHY15, JiangMYL15, JerripothulaCY16, KangOLW16} as studied by Yu \textit{et al.}~\cite{ActionProp_YuY15}.

Action proposal is a challenging task due to significant 
variations in human pose, illumination, occlusion, blur and background clutter.
In addition, the commonly available untrimmed videos bring additional noises
for accurate localization of the actions. 
Existing works have advanced the performance by
segmentation-and-merging~\cite{Jain_cvpr14,Oneata_eccv14,MaZIS13}, dense trajectories 
clustering~\cite{Jan_bmvc15}, human detection~\cite{ActionProp_YuY15} and deep learning~\cite{Li_accv16}. 
Most of these methods handle trimmed videos based on image object proposal or their variants, 
which either use sensitive low-level cues or 
separate the spatial appearance and temporal motion 
information learning into two isolated processes, 
where the later accumulates errors and also hinders the end-to-end optimization.

This paper presents a novel network fusion framework called \textit{YoTube}
that locates spatially compact and temporally smooth action proposals 
within untrimmed videos. 
A flowmap of our method is shown in Fig.~\ref{fig:flowmap}.
Specifically, our framework consists of a recurrent \textit{YoTube} detector
and a static \textit{YoTube} detector. The recurrent \textit{YoTube} explores to produce frame bounding 
boxes which exploits temporal contexts by empowering recurrent neural network (RNN) with the regression capability. 
The static \textit{YoTube} is trained to better exploit 
the rich global appearance cues within each individual frame. 
These two networks are end-to-end optimized using complementary information
of RGB and Flow, and their outputs are fused to produce accurate bounding boxes in each video frame. 
To this end, our proposed network framework overcomes the limitation of existing works by
fusing the appearance, motion and temporal context simultaneously.

The action proposals are finally constructed by linking the candidate action boxes
using dynamic programming with a novel path trimming technique. In the first pass, action paths encompassing
the whole video are generated by considering their actionness score and overlap in spatial temporal
domain. In the second pass, a novel temporal path trimming method is designed which exploits the actioness and background score transition pattern and is capable of handling the paths spanning the whole video.

Our key contributions are three folds. First, we propose 
a novel network fusion framework which learns appearance,
motion and temporal context simultaneously. 
Second, we explore the regression capability of RNN for spatial-temporal 
action proposal for the first time. Third, we design an efficient 
path trimming technique that is capable of handling
untrimmed videos without requiring time-consuming techniques of existing methods.

\section{Related Work}
\textbf{Recurrent Neural Network}:
Traditional recurrent neural network is designed to incorporate temporal dynamics 
by utilizing a hidden state in each recurrent cell. The unit works like a dynamic
memory which can be changed according to the previous states. The recurrent update
process of RNN can be modeled as follows:
\begin{equation}
\begin{split}
& h_t = \sigma(W_{xh}x_t + W_{hh}h_{t-1} + b_h)\\
& z_t = \sigma(W_{hz}h_t = b_z)
\end{split}
\end{equation}
where $h_t$ is the output hidden state and $g_t$ is the output at time $t$, and
$\sigma$ is an element-wise non-linearity. 

\begin{figure} [htb]
	\begin{center}
		\includegraphics[width = 0.35\textwidth]{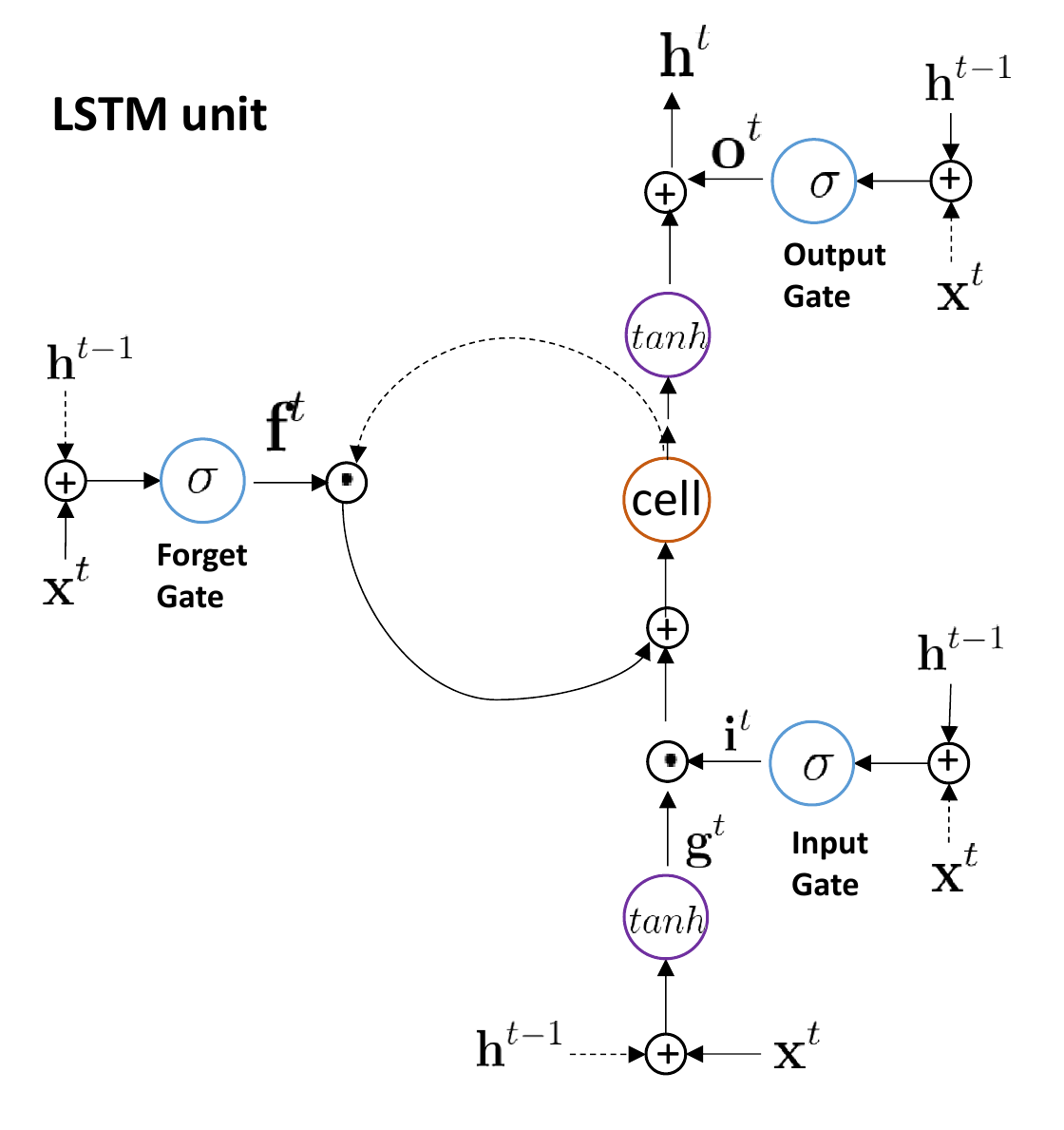}
	\end{center}
	\caption{A diagram of Long-short Term Memory \cite{HochreiterS97}.\label{fig:lstm}}
\end{figure}

However, conventional RNN is difficult to incorporate long-range
information due to the inflating or decaying of the back-propagation error over time. 
Hence Long-short Term Memory (LSTM) \cite{HochreiterS97} is proposed to 
incorporate memory that has explicitly control of 
when to 'forget' and when to 'update' given new information. 
Recently, LSTM has been actively applied to action classification 
and video description with state-of-the-art performance \cite{LRCN_DonahueHGRVDS15}. 
The LSTM unit we use in our work is similar to the one in \cite{HochreiterS97}:
\begin{equation}
\begin{split}
& i_t = \sigma(W_{xi}x_t + W_{hi}h_{t-1} + b_i)\\
& f_t = \sigma(W_{xf}x_t + W_{hf}h_{t-1} + b_f)\\
& o_t = \sigma(W_{xo}x_t + W_{ho}h_{t-1} + b_o)\\
& g_t = \sigma(W_{xc}x_t + W_{hc}h_{t-1} + b_c)\\
& c_t = f_t \odot c_{t-1} + i_t \odot g_t \\
& h_t = o_t \odot tanh(c_t)
\end{split}
\label{eq:lstm}
\end{equation}
where $i_t$ is the input gate, $f_t$ is the forget gate,
$o_t$ is the output gate, $g_t$ is the input modulation gate, and $c_t$ is the sum of previous memory cell $c_{t-1}$ which
is modulated by forget gate $f_t$ and modulation gate $g_t$. An illustration of LSTM is shown in Fig.\ref{fig:lstm}.
%%------------------------
%------------------------
%-------------------------------------------------------------------------s

%In this work, we investigate to bridge LRCN and recent popular 
%regression based object detector~\cite{YOLO_Redmon} to localize human action candidates in video 
%for the first time to the best of our knowledge.   

\textbf{Regression based Object Detection}: 
Recent advances in deep learning
have lifted the performance of state-of-the-art 
object detection. There are two paradigms: one is based on
unsupervised proposal techniques such as
selective search~\cite{SelectiveSearch_UijlingsSGS13} and
EdgeBox~\cite{EdgeBox_ZitnickD14}, which have boosted initial 
ConvNet detectors, such as Fast-RCNN. However, hand-crafted object proposal
is not directly correlated with the final recognition task which may undermine the performance of object recognition. Hence
researchers proposed to use CNN to directly regress the bounding boxes 
of object proposal using Region Proposal Network (RPN)~\cite{FasterRCNN_RenHGS15} recently. This further 
boosted the development of other regression based detection methods (e.g.
YOLO~\cite{YOLO_Redmon} and SSD~\cite{SSD_LiuAESRFB16}) that have demonstrated higher
accuracy and speed. YOLO performs inference with global image descriptor, 
hence can better exploit the context information in whole image to avoid
the influence from background. RPN and SSD use local image patches
for bounding boxes regression with faster speed, but at a cost of more bounding boxes.

However, these methods are for static images which neglects the useful 
temporal context among adjacent frames. Our work explores
to extend the regression based detectors to spatial temporal 
domain using RNN for the first time. Our study reveals 
that, RNN can capture as important information as CNN, and
their combination yields superior performance as compared with either component alone.

\textbf{Video Action Proposal}:
Early successes in action detection are based on exhaustive search using sliding cuboids~\cite{KeSH05,LanWM11,TianSS13}.
However, the rigid cuboid is difficult to capture the versatile shape
of the human actions. Besides cuboid search, Tran \textit{et al.} \cite{TranY12}
explored structure output regression to detect spatial-temporal 
action tubes, but they can only search the best action path 
with fixed size window. Although these early attempts have solved action localization problem to some extent,
the tremendously large search space leads to great computational cost.
Action proposal has great potential
to significantly reduce the search space by generating sequences of bounding boxes with good localization of candidate human actions in spatio-temporal domain. Due to the large volume of literatures,
we only review the works which are most directly related with our approach. 

Unsupervised image based object proposals \cite{SelectiveSearch_UijlingsSGS13}\cite{PrimeProposal_ManenGG13} 
have been directly extended for video action proposal.
Jain \textit{et al.}~\cite{Jain_cvpr14}
extend selective search \cite{SelectiveSearch_UijlingsSGS13} to produce action
proposal by clustering the video into voxels, which are then hierarchically
merged into action proposals. Similarly, Oneata \textit{et al.}~\cite{Oneata_eccv14} extends the work in \cite{PrimeProposal_ManenGG13}
by introducing a randomized supervoxel segmentation method for proposal generation.
Inspired by the video segmentation method by Brox
and Malik \cite{BroxM10_eccv10}, Jain \textit{et al.} \cite{Jan_bmvc15}  
propose to generation action proposals by clustering long term 
point trajectories with improved speed and accuracy. 
%Although these methods achieve good performance, their speed is relative slow
%due to the computationally expensive clustering and segmentation process. Moreover,
%these methods are sensitive to abrupt appearance and motion change in video 
%as they are based on low-level appearance and motion information which are difficult 
%to capture the underlying object information. 
%To improve the speed and accuracy of action proposal generation,  

Supervised frame-level human detection has also been introduced to further improve the performance.
Yu \textit{et al.} \cite{ActionProp_YuY15} use human detector and generate the action proposal
by using max sub-path search. Inspired by the success of deep learning, Gkioxari and Malik \cite{FindingTube_GkioxariM15}
propose to train two stream R-CNN networks \cite{RCNN_GirshickDDM14} with selective search to detect action 
regions. They link the high scored action boxes to form action tubes. 
Detection-and-tracking methods have also been used for action localization and action proposal.
Weinzaepfel \textit{et al.} \cite{WeinzaepfelHS15} train a two-stream R-CNN to detect action regions and they also train another
instance-level detectors to track the regions with Spatio-Temporal Motion Histogram. 
Li \textit{et al.} \cite{Li_accv16} train a single stream RPN~\cite{FasterRCNN_RenHGS15} to replace
R-CNN in \cite{WeinzaepfelHS15} for proposal boxes generation and use an improved method of \cite{ActionProp_YuY15} to generate
action proposal. The miss detections are remedied by tracking-by-detection and achieved the 
state-of-the-art performance. 

Although aforementioned methods have greatly advanced the quality of action proposal, they 
still have limitations. Specifically, most of these works~\cite{Jain_cvpr14,Oneata_eccv14,Jan_bmvc15,ActionProp_YuY15,FindingTube_GkioxariM15}
either produce action proposals frame-by-frame individually which ignores the interplay between appearance, motion and temporal context 
among adjacent frames or arrange the spatial information learning and temporal context learning into isolated processes
\cite{WeinzaepfelHS15, Li_accv16} which produce less satisfactory results. 
Moreover, most of these methods work on trimmed videos \cite{Jain_cvpr14,Oneata_eccv14,Jan_bmvc15,FindingTube_GkioxariM15}.
To handle untrimmed video, extra detectors need to be trained using low-level features
which further accumulates errors \cite{WeinzaepfelHS15, Li_accv16}.

Our method belongs to the two-stream deep learning based approach. Similar to these prior works, we 
train a static \textit{YoTube} detector to detect frame-level action candidates, but we perform reasoning using global image features to reduce the interference from background clutter. In addition, we trained a recurrent \textit{YoTube} detector to capture the long-term dependency and context among adjacent frames which are largely neglected in those existing methods. The two networks can be optimized end-to-end by simultaneously integrating appearance, motion and temporal dynamics, and their outputs are further fused for accurate candidate action box prediction.
Moreover, we design a novel and efficient path trimming technique which is capable of handling untrimmed videos directly without requiring time-consuming techniques of existing methods

%%%%%%%%%%%%%%%%%%%%%%%%%%%%%%%%%%%%%%%%%%%%%%%%%%%%%%%%%%%%%%%%%%%%%%%%%%%%
\section{Methodology}
The proposed method takes an untrimmed video as input and outputs the action proposal accordingly as illustrated in Fig.~\ref{fig:flowmap}. 
Our framework consists of two steps: (1) sequential candidate action bounding boxes prediction using recurrent and static \textit{YoTube}; (2) action path linking and trimming.  Further details of our method are elaborated in the following sections.

%%------------------------
\begin{figure} [htb]
	\begin{center}
		\includegraphics[width = 0.5\textwidth]{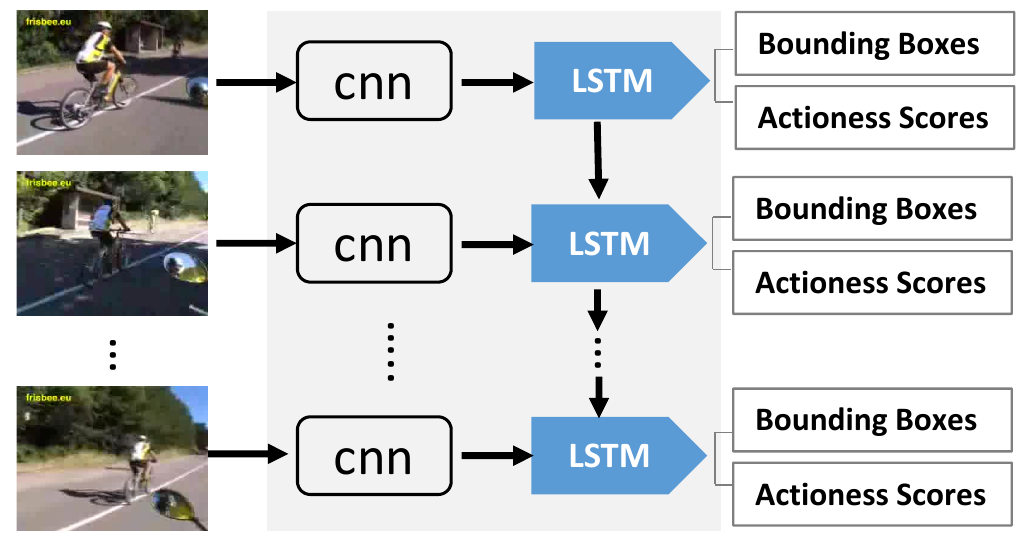}
	\end{center}
	\caption{\label{fig:lstm_yolo} With a video snippet as input, the proposed recurrent YoTube detector first extracts discriminative features from each frame and then apply the LSTM to regress the coordinates of the bounding boxes. The bounding boxes of each frame is estimated by considering the rich spatial and temporal context in forward direction.}
\end{figure}
\vspace{-15pt}
%------------------------
%-------------------------------------------------------------------------s
\subsection{YoTube for Action Candidate Boxes Generation}
\label{sec:yolo}
One limitation of existing deep learning based action proposals methods is they either process a video frame-by-frame~\cite{FindingTube_GkioxariM15} 
or separate spatial and temporal information learning into two isolated processes \cite{Li_accv16, WeinzaepfelHS15}. On the 
other hand, temporal dynamics and contexts among adjacent frames have been proven to be useful for recent state-of-the-arts in 
action classification and video description~\cite{LRCN_DonahueHGRVDS15}. Inspired by this idea, we 
design a network fusion framework which can incorporate apperance, motion and temporal context learning 
in a unified and end-to-end optimizable manner. 

We firstly describe the \textit{\textbf{recurrent YoTube}}, which is a Long-term Recurrent Convolutional Network (LRCN) \cite{LRCN_DonahueHGRVDS15} that combines CNN and RNN for sequence bounding boxes prediction. 
Note LRCN was initially designed for action classification, which maps a sequence of input 
feature vectors into sequence of frame lables. In this work, we adapt LRCN to map the sequence
of feature vectors to sequence of tensors encoding the object bounding boxes information,
as inspired by the recent popular regression based object detectors \cite{FasterRCNN_RenHGS15, YOLO_Redmon}.

Fig. \ref{fig:lstm_yolo} depicts the architecture of recurrent \textit{YoTube} which
works by passing each video frame $f_t$ at time $t$ into a CNN to produce a fixed-length
feature $x_t$. Then $x_t$ is passed into the recurrent LSTM, 
which maps the input $x_t$ and previous time step
hidden state $h_{t-1}$ to a new hidden state $h_t$ and bounding boxes $o_t$ as in Eq. \ref{eq:lstm}. The inference is conducted sequentially from top to bottom as illustrated in Fig. \ref{fig:lstm_yolo},
%where $h_1 = f_W(x_1, h_0)$, then $h_2 = f_W(x_2, h_1)$, etc to $h_T$. The notation $W$ is the weight
%of LSTM, 
hence the context in earlier frames $t_l$ ($t_l < t$) can be propagated to the current frame $t$.

The output $o_t$ for frame $t$ is a $K\times K\times(B\times 5 + |S|)$ tensor encodes the output bounding boxes information. Specifically, it means to
divide the image into $K\times K$
grids. Each grid cell will predict 
$B$ bounding boxes which are parameterized by $(x, y, w, h, c)$
where $(x, y)$ represents the center of the box relative
to the bounds of the cell. The width $w$ (or height $h$) is
normalized with respect to the image width (height). The confidence $c$
predicts the IoU between the predicted box and any
ground-truth box. Moreover, each cell will also predict a score tuple $S = (s_{ac}, s_{bg})$, 
where $s_{ac}$ and $s_{bg}$ is an actionness score 
and a background score for the given cell, respectively.

The loss function to be minimized is defined as a sum-squared error between prediction $o_t$ and ground-truth $\hat o_t$ for optimization simplicity \cite{YOLO_Redmon}:
\begin{equation}
\begin{split}
&\lambda_{coord}\sum_{i=0}^{K^2}\sum_{j=0}^{B}1_{ij}^{obj}\lVert (x_i, y_i) -(\hat x_i, \hat y_i)\rVert^2\\
&+\lambda_{coord}\sum_{i=0}^{K^2}\sum_{j=0}^{B}1_{ij}^{obj}\lVert (\sqrt{h_i}, \sqrt{w_i}) -(\sqrt{\hat h_i}, \sqrt{\hat w_i})\rVert^2\\
&+\sum_{i=0}^{K^2}\sum_{j=0}^{B}1_{ij}^{obj}(c_i - \hat c_i)^2\\
&+\lambda_{noobj}\sum_{i=0}^{K^2}\sum_{j=0}^{B}1_{ij}^{noobj}(c_i - \hat c_i)^2\\
&+\sum_{i=0}^{K^2}1_{i}^{obj}\sum_{k\in\{ac,bg\}}(s_k^i-\hat s_k^i)^2
\end{split}
\label{eq:objective}
\end{equation} where $\hat o_t^i = (\hat x_i, \hat y_i, \hat h_i, \hat w_i, \hat c_i, \hat s_{ac}^i, \hat s_{bg}^i)$ is the cell $i$ of the ground-truth $\hat o_t$, 
$1^{obj}_i$ denotes if object appears in cell $i$, $1^{obj}_{ij}$ denotes that the $j^{th}$ bounding box predictor in cell $i$ is responsible for the prediction (i.e. has the higher IoU with the ground truth between the $B$ boxes) and $1^{noobj}_{ij}$ denotes that the $j^{th}$ bounding box predictor in cell $i$ is not responsible for the prediction or that there is no ground truth boxes in cell $i$. 

The first two terms penalize coordinates error only when the prediction is responsible for the ground truth box. As deviation in the predicted coordinates matters more for small boxes than large boxes, we take the square root of width and height. The third and fourth terms penalize confidence score error, reduced by a factor $\lambda_{noobj}$ when the prediction is not responsible for the ground truth box. As most of the grid cells don't contain object, it limits to push confidence score towards zero. The final term penalizes classification as "action" or "background" error only when there is an object in the cell. In this work, we set $\lambda_{coord} = 5$ and $\lambda_{noobj} = 0.5$.

Recurrent \textit{YoTube} is doubly deep in spatial-temporal domain, which can learn the temporal action dynamics.
To further exploit the rich RGB and Flow cues in individual frame, we train a　~\textit{\textbf{static YoTube}} 
which shares the same architecture as the recurrent \textit{YoTube}, but replaces the last LSTM layer with a fully-connected layer of same number of neurons to regress the coordinates. These two networks complement each other and their outputs
are combined to further improve the performance. 

%\subsubsection{Inference}
%
%During test time, we predict 98 boxes per frames. To reduce the computational complexity of the tube generation, we discard some boxes given their size ($< 10\times 10$ pixels) and their confidence score after rescoring ($<0.1$ or less if we want more boxes).
%
%We rescore the confidence score to rescale it between 0 and 1 where 1 will be the score of the highest scored box and 0 the one of the smallest scored box. This rescaling permits a fairer comparison of successive frames from the same video.

\subsection{Path Linking and Trimming}
\label{sec:action_tube}
At the end of the detection process (Sec. \ref{sec:yolo}), we have a set of bounding boxes for each frame of the video $\mathbf{B} = \{\{b_i^{(j)}, j\in [1\ldots N_{b_i}]\}, i \in [1\ldots T]\}$ where $T$ is the length of the video and $N_{b_i}$ is the number of predicted boxes in frame $i$. For each box $b_i^{(j)}$ we have its confidence score $s_c(b_i^{(j)})$, actionness score $s_{ac}(b_i^{(j)})$ and background score $s_{bg}(b_i^{(j)})$. The next objective is to create a set of proposal paths $\mathbf{P} = \{ p_i = \{b_{m_i}, b_{m_{i+1}}\ldots b_{n_i}\}, i\in [1\ldots |\mathbf{P}|]\}$ where $m_i$ and $n_i$ are the starting and ending frame of path $p_i$, respectively.

\subsubsection{Action Path Linking}
\label{sec:path_linking}
In order to link frame-level boxes into coherent path, we firstly define a score for each path given its
 confidence scores $s_c$ of each box and the IoU of successive boxes:
\begin{equation}
S(p) = \underbrace{\sum\limits_{i=1}^T s_c(b_i)}_{unary} + \lambda_0\times\underbrace{\sum\limits_{i=2}^T IoU(b_i,b_{i-1})}_{pairwise}
\label{eq:dynamic}
\end{equation}
$S(p)$ is high for path whose detection boxes have high confidence scores and overlap significantly. $\lambda_0$ is a trade-off factor 
to balance the two terms. 

Maximizing Eqn.~\ref{eq:dynamic} helps find paths whose
detection box scores are high and consecutive detection boxes overlap significantly
in spatial and temporal domain. The path which
maximizes the energy $\hat p_c = \underset{p_c}{argmax} E(p_c)$ can be found with the Viterbi algorithm \cite{FindingTube_GkioxariM15}.
Once an optimal path has been found, we remove the bounding boxes in previous path from the frames
to construct next path until certain frame doesn't have any boxes. 

\begin{figure}[htb]
	\centering
    \includegraphics[width = 0.4\textwidth]{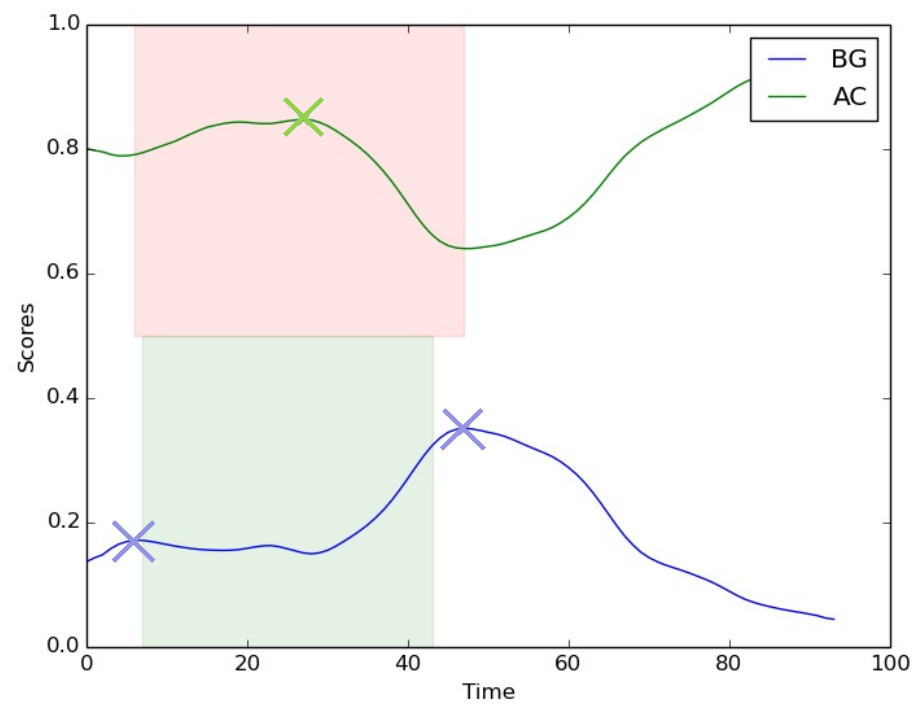}
	\caption{Illustration of the proposed peak trimming method on one UCF-101 videos: Blue and green curves represent the background and actionness scores, respectively, where blue and green crosses denote score peaks. Green patches represent the ground-truth paths and red patches represent paths that are extracted by using the proposed peak trimming method. It can be observed that the proposed method is capable of trimming the predicted paths accurately thanks to certain action and background score transition patterns.}
	\label{peaks}
\end{figure}
\subsubsection{Action Paths Trimming}
\label{sec:path_trimming}
The generated action paths as described in the last subsection spans the entire video as it 
greedily optimizes all confidence scores across the paths. On the other hand, human 
actions typically take up a fraction of it for untrimmed video. It is therefore necessary
to perform trimming to remove those boxes that are unlikely belong to the action regions. 
Mathmetically, we would like to assign each box $b_t$ in a path $p$ with
a binary label $y_t \in \{0, 1\}$ (where 'zero' and 'one' represent the 'background' and 'action' class respectively), such that the boxes which are near (or far) from the valid action regions should be assigned to 'action' (or 'background') class as much as possible in final path labeling $\hat Y_p = [\hat y_0, \hat y_1, ..., \hat y_T]$. 

We also noticed that a transition in background scores typically signifies a change between action and non-action frames. In addition, the boxes within a valid action region often have high actioness scores. Hence, detecting peaks in actioness scores helps find a potential action region, while finding peaks in background score helps define the start and end of the action regions.

Hence, we propose a new method by looking at the transition pattern in the actionness and background 
scores for the path trimming as illustrated in Fig.\ref{peaks}. We first smooth the scores by computing their running average to reduce the influence of noisy classifier scores. 
All the peaks in both scores are then detected where a peak is defined as a local maximum among at least $n$ neighbors:
%\begin{gather*}
%peaks_{ac} = \{t, s_{ac}(b_t) = \max(V_n^{(ac)}(t))\}\\
%peaks_{bg} = \{t, s_{bg}(b_t) = \max(V_n^{(bg)}(t))\}
%\end{gather*}
\begin{equation}
\begin{split}
peaks_{ac} = &\{t, s_{ac}(b_t) = \max(V_n^{(ac)}(t))\}\\
peaks_{bg} = &\{t, s_{bg}(b_t) = \max(V_n^{(bg)}(t))\}
\end{split}
\end{equation}
where $V_n^{(k)}(t) = \{s_{k}(b_i), i\in [t-n\ldots t+n]\}, k\in\{ac, bg\}$.

Once we have found the peaks, we can select all subsequences to generate the final action proposals by applying the following algorithm:
\begin{algorithm}[htb]\label{eq:peak_trim}
	\caption{Action paths trimming using actioness and background score peaks.}
	\begin{algorithmic}
	\State \textbf{Input}: actioness score peaks $peaks_{ac}$ and background score peaks $peaks_{bg}$.
	\State \textbf{Output}: set $subseq$ consists of trimmed paths.\\
	
	\State $subseq = \emptyset$ 
	\For{$p\in peaks_{ac}$}
		\State $s = \max(peaks_{bg} < p)$ 
		\State $e = \min(peaks_{bg} > p)$ 
		\State add path $\{b_s\ldots b_e\}$ to $subseq$ 
    \EndFor
	\end{algorithmic}
\end{algorithm}

%\begin{algorithm}[htb]
%	$subseq = \emptyset$\\
%	\For{$p\in peaks_{ac}$}{
%		$s = \max(peaks_{bg} < p)$\\
%		$e = \min(peaks_{bg} > p)$\\
%		add path $\{b_s\ldots b_e\}$ at $subseq$
%	}
%\end{algorithm} \label{eq:peak_trim}

%Inspired by these observations, we design a novel objective function that encourages the selected paths' labelings to have a high 'actioness' scores and at the same time penalizes the inconsistent labeling among boxes of adjacent frames as follows:
%\begin{equation}
%\label{eq_trimming}
%\hat Y_p = \underset{Y_p}{argmax} \sum_{t=1}^{T}s(y_t) - \lambda_1\times\sum_{t=2}^{T}g(y_t, y_{t-1})
%\end{equation}
%where $\lambda_1$ is a trade-off parameter, $s(y_t) = s_{ac}(b_t)$ (or $s_{bg}(b_t)$)
%when $y_t$ is assigned to 'one' (or 'zero'), and pairwise potential $g$  penalizes the 
%inconsistent labeling between $y_t$ and $y_{t-1}$ which is defined to be: 
%\begin{equation}
%\label{Potts}
%g(y_t, y_{t-1}) = \left\{
%\begin{array}{ll}
%0 & \mbox{if } y_t = y_{t-1} \\
%\alpha_{y_t} & \mbox{otherwise}
%\end{array}
%\right.
%\end{equation}
%where the penalty $\alpha_{y_t}$ is set by cross-validation. 
%Eqn.~\ref{eq_trimming} can also be solved by the Viterbi algorithm as in previous step.

%==============================================================================
\section{Implementation and Benchmarking}
In this section, we discuss the details of implementation and benchmarking,
including the dataset and evaluation metrics.
%%==========================================================================
\begin{figure*}
	\centering
	\includegraphics[width = 1\textwidth]{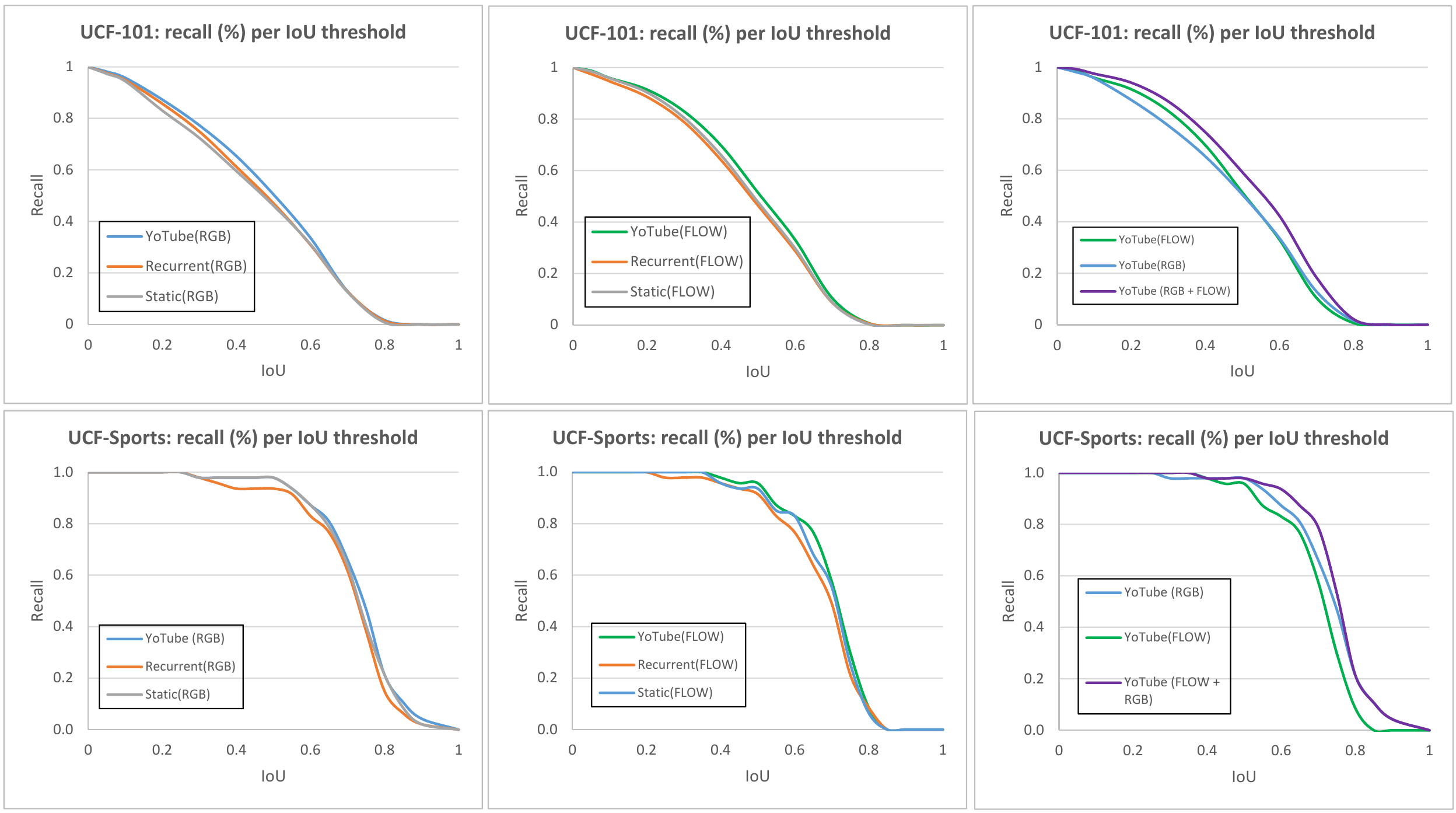}
	\caption{Ablation study of two stream's static YoTube and recurrent YoTube on the UCF-101 (top-row) and UCF-Sports (bottom row) datasets, left column shows RGB stream, middle column shows the Flow stream and right column shows their ensemble.}
	\label{fig:recall_vs_iou_ablation}
\end{figure*}
%%==========================================================================

\subsection{Training}
The CNN architecture we use for feature extraction in \textit{YoTube} is adapted from \cite{YOLO_Redmon} which has 24 convolution layers and 2 fully-connected layer. We firstly 
replacing the last two fully connected layer with a locally connected layer with 256 filters with a $3\times3$ kernel. On top of the locally connected layer, we train an LSTM layer with 588 neurons which directly regresses the bounding boxes' coordinates. We choose locally connected layer to stablize training and improve convergence. 
The number of neurons in last layer means to divide the image into $7\times7$ grids and each grid predicts $2$ bounding boxes.

For the RGB stream, the convolutional part of our model is pretrained on the ImageNet 1000-class dataset \cite{ILSVRC15}. For the Flow stream, the convolutional part is pretrained with the weights of the RGB stream. The top layers are initialized using the method in \cite{HeICCV15}. We found no problem in convergence by initializing the weights of our Flow model with the weights of the RGB model despite the notable difference between the images distribution.

We make an extensive use of data augmentation to prevent over-fitting. This part is non negligible due to important correlation between frames of the same video. In addition to mirroring, we use corner cropping and center cropping. It means that we take a $224\times 224$ crop from the $320\times 240$ frame in each corner and the center. Then we resize this crop to the input size of $448\times 448$. This method permits to increase the size of the dataset by a factor 12.

We use Adam \cite{Adam} optimizer during training with default parameters. When training the static \textit{YoTube}, we use a batch size of 32 frames from different videos during 100 epochs with an initial learning rate of $10^{-4}$ decaying at $10^{-5}$ after the $20^{th}$ epoch. When training the recurrent \textit{YoTube}, we freeze the weighs of the convolutional layers to avoid a catastrophic forgetting. We then use a batch size of 10 sequences of 10 frames from different videos during 50 epochs. The same learning rate planning is used.

\subsection{Datasets}
\label{sec:dataset}

\textbf{UCF-101.} The UCF-101 dataset is a large action recognition dataset
containing 101 action categories with more than 13,000 videos and an average
of 180 frames per video. A subset of 24 categories are used for the localization task
with bounding box annotation corresponding to 3,204 videos. Part of these videos
are untrimmed (around 25$\%$), permitting to validate the efficiency of our trimming methods.
Each video contains one or more instances of same
action class. It has large variations in terms of appearance, scale, motion, etc with 
much diversity in terms of actions. Three train/test splits are provided with
the dataset, and we perform experiments on the first split with 2,290 training videos and
914 testing videos. 

\textbf{UCF-Sports.} This dataset contains 150 sport broadcast videos 
with realistic actions captured in dynamic and cluttered environments. 
It is challenging considering many actions with large 
displacement and intra-class variation. These videos have been 
trimmed to contain a single action instance without interruption.
There are ten categories in the dataset, e.g. ‘diving’, ‘swinging
bench’, ‘horse riding’, etc. We used the train-test split of videos suggested in \cite{Jan_bmvc15} (103 for training and 47 for testing). The ground truth is provided as sequences
of bounding boxes enclosing the actions. 

\subsection{Evaluation metrics}
\label{sec:metrics}
\textbf{ABO, MABO:} We use two common metrics as in \cite{Jan_bmvc15} to report overall performance, namely Average Best Overlap (ABO) and Mean ABO (MABO).
The overlap (OV) between a path $\mathbf{d} = \{d_s\ldots d_e\}$ and a ground truth path $\mathbf{g} = \{g_s\ldots g_e\}$ is defined as follows:
\begin{align*}
	OV(\mathbf{d},\mathbf{g}) & = \frac{1}{|\mathbf{d}\bigcup\mathbf{g}|}\times\sum\limits_{i\in \mathbf{d}\bigcap\mathbf{g}}\frac{d_i\bigcap g_i}{d_i\bigcup g_i}\\
	|\mathbf{d}\bigcup\mathbf{g}| & = \max(d_e, g_e) - \min(d_s, g_s)\\
	\mathbf{d}\bigcap\mathbf{g} & = [\max(d_s,g_s)\ldots \min(d_e, g_e)]
\end{align*}
where $d_s$ and $d_e$ are the detected bounding boxes in the starting and ending frame of a path, $g_s$ and $g_e$ are the bounding boxes in the starting
and ending frame of the ground-truth path.

ABO measures the best localization from the set of action proposals $D = \{d_j | j= 1...m\}$ for the each ground-truth $G$, where ABO(c) is the ABO computed for the ground-truth $G_c$ of class $c$. The mean ABO (MABO) measures the average performance across all classes.
\begin{align*}\label{metric}
\mbox{ABO} & = \frac{1}{|\mathbf{G}|}\sum\limits_{\mathbf{g}\in \mathbf{G}} \max_{d\in\mathbf{D}} OV(\mathbf{d},\mathbf{g})\\
\mbox{ABO}(c) & = \frac{1}{|\mathbf{G^c}|}\sum\limits_{\mathbf{g}\in \mathbf{G^c}} \max_{d\in\mathbf{D}} OV(\mathbf{d},\mathbf{g})\\
\mbox{MABO} & = \frac{1}{|\mathbf{C}|}\sum\limits_{c\in\mathbf{C}}\mbox{ABO}(c)
\end{align*}
where $\mathbf{C}$ is the set of action classes, $\mathbf{G}$ is the set of ground truth paths and $\mathbf{G}^c$ is the set of ground truth paths for action class $c$. 

\textbf{Recall vs IoU:} Another commonly used metric is the Recall vs IoU~\cite{EdgeBox_ZitnickD14}, which measures the fraction
of ground-truths detected in a set of overlap threshold. An instance of action, $g_i$ is correctly detected by an action proposal $d_j$ if the overlap score is higher than a threshold $\eta$ \textit{i.e.}: $OV(d_j, g_i) \geq \eta$ where $\eta \in [0, 1]$.
In our work, we target to maximize the recall at a $0.5$ threshold as other works~\cite{Jan_bmvc15, Li_accv16}.
%%==========================================================================
\begin{figure*}
	\centering
	\includegraphics[width = .95\textwidth]{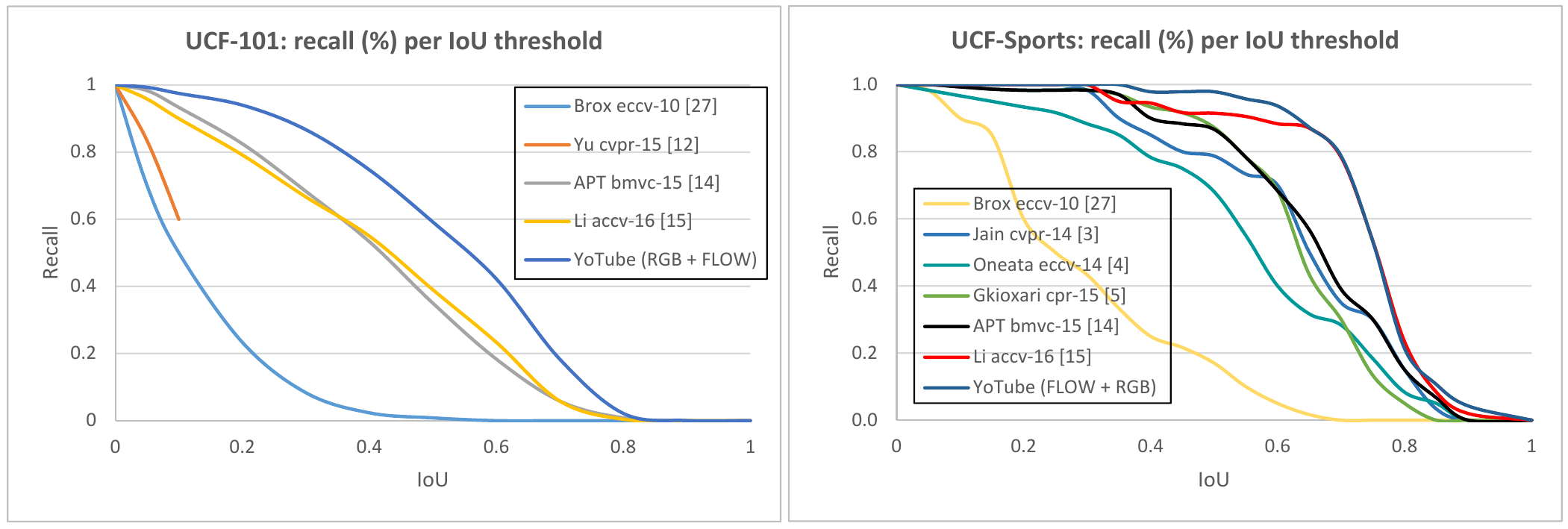}
	\caption{Comparison with other state-of-the-arts on UCF-101 and UCF-Sports dataset, performance is measured by recall for various IoU thresholds.}
	\label{fig:recall_vs_iou_overall}
\end{figure*}
%%==========================================================================
\begin{figure*}
	\centering
	\includegraphics[width = .95\textwidth]{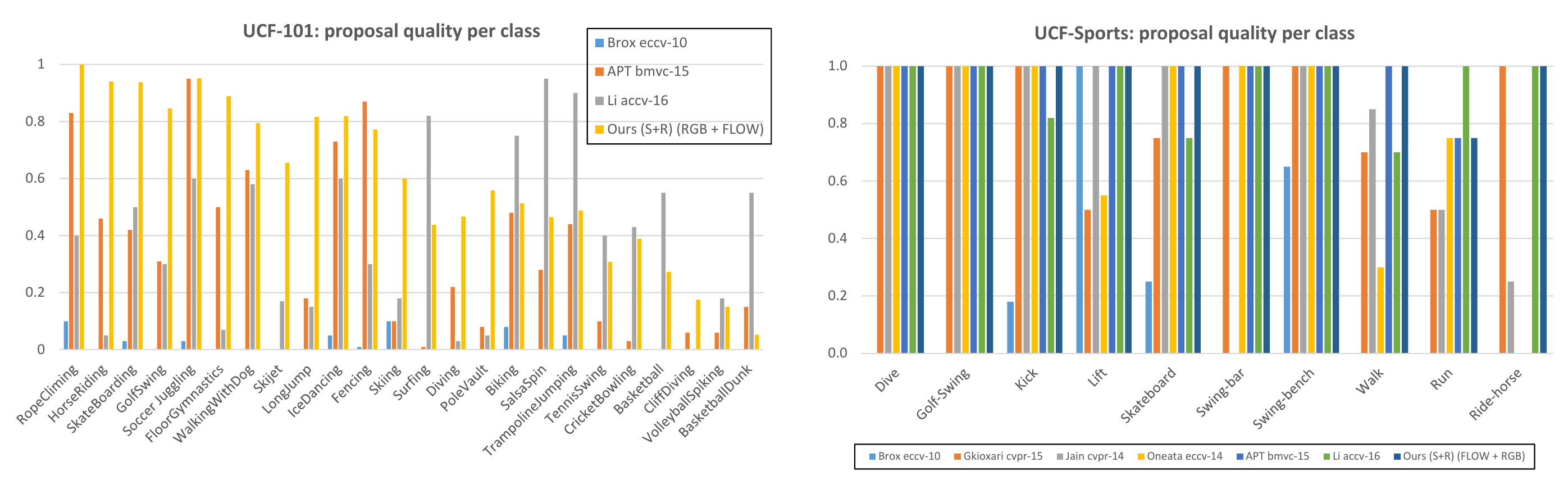}
	\caption{Comparison with other state-of-the-arts on UCF-101 and UCF-Sports, performance is measured by the recall on each action class.}
	\label{fig:recall_per_class_overall}
\end{figure*}
%%==========================================================================
%%%%%%%%%%%%%%%%%%%%%%%%%%%%%%%%%%%%%%%%%%%%%%%%%%%%%%%%%%%%%%%%%%%%%%%%%%%%
\section{Experimental Results}
In our experiments, we first compare recurrent \textit{YoTube} and static \textit{YoTube} in rgb and optical flow streams. Then, we study the impact of path-trimming on the detecction performance. Finally, we compare the \textit{YoTube} with the state-of-the-art.

\begin{table}
	\centering
	\begin{tabular}{|l|c|c|c|c|}
		\hline
		UCF101 & ABO & MABO & Recall & \#Prop. \\
		\hline
		RGB Stream & ~ & ~ & ~ & ~ \\
		\hline
		Static (S) & 44.94 & 45.42& 46.13& \textbf{10} \\
		Recurrent (R) & 45.85 & 45.83& 47.05& 21\\
		YoTube (RGB) & \textbf{47.60} & \textbf{47.78}& \textbf{50.61}& 35\\
		\hline
		Flow Stream & ~ & ~ & ~ & ~ \\
		\hline
		Static (S) & 46.87 & 47.02& 47.63& 33 \\
		Recurrent (R) & 46.09 & 46.45 & 46.3& \textbf{9}\\
		YoTube (FLOW) & \textbf{48.61} & \textbf{48.83} & \textbf{51.29}& 42\\
		\hline
		Ensemble & ~ & ~ & ~ & ~ \\
		\hline	
		YoTube-RGB+FLOW (NO TRIM) & 45.36& 46.42& 52.03 & 80\\	
		YoTube-RGB+FLOW & \textbf{52.45}& \textbf{52.92}& \textbf{59.19} & 73\\
		\hline
	\end{tabular}
	\caption{\label{table:ablation_ucf101} Ablation study on the UCF-101 dataset. }
\end{table}

\begin{table}
	\centering
	\begin{tabular}{|l|c|c|c|c|}
		\hline
		UCF-Sports & ABO & MABO & Recall & \#Prop. \\
		\hline
		RGB Stream & ~ & ~ & ~ & ~ \\
		\hline
		Static (S) & 71.64& 72.54& \textbf{97.87}& \textbf{20} \\
		Recurrent (R) & 70.08& 71.5& 93.62& \textbf{20}\\
		YoTube (RGB) & \textbf{72.45}& \textbf{73.54}& \textbf{97.87}& 30\\
		\hline
		Flow Stream & ~ & ~ & ~ & ~ \\
		\hline
		Static (S) & 68.08& 68.98& 93.62& \textbf{20}\\
		Recurrent (R) & 66.22& 66.91& 91.49& \textbf{20}\\
		YoTube (FLOW) & \textbf{69.08}& \textbf{69.95}& \textbf{95.74}& 30\\
		\hline
		Ensemble & ~ & ~ & ~ & ~ \\
		\hline
		YoTube-RGB+FLOW (NO TRIM)& \textbf{74.44}& \textbf{75.31}& \textbf{97.87}& 30\\		
		YoTube-RGB+FLOW& \textbf{74.44}& \textbf{75.31}& \textbf{97.87}& 30\\
		\hline
	\end{tabular}
	\caption{\label{table:ablation_ucfsport} Ablation study on the UCF-Sports dataset. }
\end{table}

\subsection{Recurrent YoTube vs Static YoTube}
The ablation comparison between recurrent and static \textit{YoTube} for two streams in UCF-101 and UCF-Sports are shown in Fig.~\ref{fig:recall_vs_iou_ablation}, Table \ref{table:ablation_ucf101} and \ref{table:ablation_ucfsport}. In UCF-101, the performance of recurrent version 
is slightly better than the static version in RGB stream 
with around $1\%$ improvements in recall and 
the ensemble model (YoTube(RGB)) achieves another $3.56\%$ improvement as shown in Table \ref{table:ablation_ucf101}. These results prove that the recurrent model and static model are complementary.  
We conjecture this is due to the RNN which captures the temporal dynamics among adjacent frames. 
For the results of Flow stream in Table \ref{table:ablation_ucf101}, the recall of static version is $1.3\%$ better than the recurrent version and the ensemble model (YoTube(FLOW)) achieves another $4\%$ improvement than the static version, which further confirms the complementariness of two methods. The slightly inferior result of recurrent version is probably caused by lacking training data. In addition, the ensemble flow stream (YoTube(FLOW)) performs slightly better than the ensemble rgb stream (YoTube(RGB)). This is probably because the flow field eliminates the interference from the background. The final model (YoTube (RGB+FLOW)) which combines two streams achieves another $8\%$ improvement than the flow stream in recall, which demonstrates that the RGB and Flow streams also complement each other. 

For UCF-Sports dataset, the static version has $4\%$ better recall than the recurrent version as shown in Table \ref{table:ablation_ucfsport}, whereas the ensemble model have the same recall as the static version, it has higher ABO and MABO for better localization. For the flow stream, the static
version has $2.2\%$ better recall than the recurrent version in the interval, the ensemble model is $2\%$ better recall than the static version. The slightly inferior performance of recurrent YoTube could be caused by the small number of training sample. On the other hand, the higher ABO and MABO of the ensembles between static and recurrent model in both streams still proves that the two models capture complementary information to improve localization. 

Moreover, we also show the result of the model without path trimmming (NO TRIM) in both Table \ref{table:ablation_ucf101} and \ref{table:ablation_ucfsport}. As UCF-101 dataset contains un-trimmed video, the performance of our method without path trimming is nearly $7\%$ lower in recall and also contains more noisy paths. This shows that the proposed path trimming is effective for the untrimmed video. For the UCF-Sports dataset which consists of trimmed videos, the performance with or without path trimming is identical. This result also proves that our method is adaptive to the video content. 

\subsection{Comparison to state-of-the-arts}
We compare our method with state-of-the-arts on UCF-101
and UCF-Sports datasets. The recall-vs-IoU and recall-per-class curves 
for both datasets are shown in Fig.~\ref{fig:recall_vs_iou_overall}
and Fig.~\ref{fig:recall_per_class_overall}. 

For UCF-101 dataset, our method out-performs the state-of-the-art~\cite{Li_accv16} by $20\%$ or more in all range of IoU.
Although Li \textit{et al.}~ use deep network (RPN)~\cite{FasterRCNN_RenHGS15}, 
they use only one stream and their performance is only $4\%$ better than the unsupervised method APT \cite{Jan_bmvc15} in terms of the recall as shown in Table ~\ref{table:ucf101}. 
Notwithstanding, the recall of our single stream design in RGB and Flow out-performs Li \textit{et al.}\cite{Li_accv16}
by $11\%$ and $12\%$ respectively, which proves the superiority of \textit{YoTube} by using spatial-temporal modeling and two-stream design. Yu \textit{et al.}~\cite{ActionProp_YuY15} is based on a human detector with low-level features, which 
is difficult to handle large dynamic changes in the scene.
The work in \cite{BroxM10_eccv10} is mainly designed for non-overlap segmentation and using 
low-level features, hence it is sub-optimal for the task and has the lowest recall. 
According to the per-class recall curve in Fig.~\ref{fig:recall_per_class_overall}, our method 
is better than Li \textit{et al.}~\cite{Li_accv16} in many classes, especially for classes 
that have large motion change (e.g. 'skijet', 'floor gymnastics' and 'long jump' ).

For UCF-Sports dataset, our method also out-performs Li \textit{et al.} \cite{Li_accv16} by nearly $6\%$ in terms of recall in Table \ref{table:ucfsports}. Moreover, 
the deep learning based approaches (our method and Li \textit{et al.}~\cite{Li_accv16}) also
largely out-perform the unsupervised methods, which proves the effectiveness of the deep networks' discriminative features. The per-class recall curve for all methods is also provided in Fig.~\ref{fig:recall_per_class_overall}. 

We also evaluate our methods in terms of other metrics, e.g. ABO, MABO, number of proposals 
which are also listed in Table \ref{table:ucf101} and \ref{table:ucfsports}. Our method produces
the highest MABO, Recall using slighly higher number of proposal than Li \textit{et al.} \cite{Li_accv16},
while still relatively smaller than \cite{Jan_bmvc15} and \cite{ActionProp_YuY15}. Although  
\cite{Li_accv16} achieves the highest ABO in both datasets, 
there are big differences between ABO and MABO. 
Actually, these two measurements should be in the similar scale according to the formula
in Sec.~\ref{sec:metrics}, i.e. MABO is the mean of ABO for all classes. 

%Some visual examples are shown in Fig.\ref{predicted-ucf101} and \ref{predicted-ucfsports} which demonstrates that 
%our method can produce candidate paths which have good localization of the human actions.  

\begin{table}
	\centering
	\begin{tabular}{|c|c|c|c|c|}
%		\hline
%		~ & \multicolumn{4}{c|}{UCF101} \\
		\hline
		UCF101 & ABO & MABO & Recall & \#Prop. \\
		\hline
		Brox \& Malik~\cite{BroxM10_eccv10} & 13.28 & 12.82 & 1.40 & \textbf{3} \\
		Yu \textit{et al.}~\cite{ActionProp_YuY15} & n.a & n.a & 0.0 & 10,000\\
		APT~\cite{Jan_bmvc15} & 40.77 & 39.97 & 35.45 & 2299 \\
		Li \textit{et al.}~\cite{Li_accv16} & \textbf{63.76} & 40.84 & 39.64 & 18 \\
		\hline
		YoTube-RGB& 47.60& 47.78& 50.61& 35 \\
		YoTube-FLOW& 48.61& 48.83& 51.29& 42\\
		YoTube-RGB+FLOW& 52.45& \textbf{52.92}& \textbf{59.19}& 73\\
		\hline
	\end{tabular}
	\caption{\label{table:ucf101} Quantitative comparison on the UCF-101 dataset. Recall is computed at a precision threshold of $0.5$.}
\end{table}
\begin{table}
	\centering
	\begin{tabular}{|c|c|c|c|c|}
		\hline
%		~ & \multicolumn{4}{c|}{UCF Sports} \\
		UCF Sports & ABO & MABO & Recall & \#Prop. \\
		\hline
		Brox \& Malik~\cite{BroxM10_eccv10} & 29.84 & 30.90 & 17.02 & \textbf{4} \\
		Jain~\textit{et al.}~\cite{Jain_cvpr14} & 63.41 & 62.71 & 78.72 & 1642 \\
		Oneata~\textit{et al.}~\cite{Oneata_eccv14} & 56.49 & 55.58 & 68.09 & 3000\\
		Gkioxari~\textit{et al.}~\cite{FindingTube_GkioxariM15} & 63.07 & 62.09 & 87.23 & 100 \\
		APT~\cite{Jan_bmvc15} & 65.73 & 64.21 & 89.36 & 1449 \\
		Li \textit{et al.}~\cite{Li_accv16} & \textbf{89.64} & 74.19 & 91.49 & 12 \\
		\hline
		YoTube-RGB& 72.45& 73.54& 97.87& 30\\
		YoTube-Flow& 69.08& 69.95& 95.74& 30\\
		YoTube-RGB+Flow& 74.44& \textbf{75.31}& \textbf{97.87}& 30\\
		\hline
	\end{tabular}
	\caption{\label{table:ucfsports} Quantitative comparison on the UCF-Sports dataset. Recall is computed at a precision threshold of $0.5$.}
\end{table}

\section{Conclusion}
We propose a novel framework for video action proposal. Given an untrimmed
video as input, our method produces a small number of spatially compact and temporally 
smooth action proposals. The proposed method explores the 
regression capability of RNN and CNN 
to produce frame-level candidate action boxes using rgb, flow and temporal contexts 
among frames, hence greatly increases the accuracy and meanwhile 
reduces the number of false positives. The action proposals are constructed using dynamic with a novel path trimming methods. The experiments on the UCF-Sports and UCF-101 datasets highlight the effectiveness of our proposed modeling.



\bibliographystyle{IEEEtran}
\bibliography{egbib2}

\begin{thebibliography}{10}
\providecommand{\url}[1]{#1}
\csname url@samestyle\endcsname
\providecommand{\newblock}{\relax}
\providecommand{\bibinfo}[2]{#2}
\providecommand{\BIBentrySTDinterwordspacing}{\spaceskip=0pt\relax}
\providecommand{\BIBentryALTinterwordstretchfactor}{4}
\providecommand{\BIBentryALTinterwordspacing}{\spaceskip=\fontdimen2\font plus
\BIBentryALTinterwordstretchfactor\fontdimen3\font minus
  \fontdimen4\font\relax}
\providecommand{\BIBforeignlanguage}[2]{{%
\expandafter\ifx\csname l@#1\endcsname\relax
\typeout{** WARNING: IEEEtran.bst: No hyphenation pattern has been}%
\typeout{** loaded for the language `#1'. Using the pattern for}%
\typeout{** the default language instead.}%
\else
\language=\csname l@#1\endcsname
\fi
#2}}
\providecommand{\BIBdecl}{\relax}
\BIBdecl

\bibitem{LRCN_DonahueHGRVDS15}
J.~Donahue, L.~A. Hendricks, S.~Guadarrama, M.~Rohrbach, S.~Venugopalan,
  T.~Darrell, and K.~Saenko, ``Long-term recurrent convolutional networks for
  visual recognition and description,'' in \emph{CVPR}, 2015.

\bibitem{2Stream_WangSWVH16}
Y.~Wang, J.~Song, L.~Wang, L.~{Van Gool}, and O.~Hilliges, ``Two-stream sr-cnns
  for action recognition in videos,'' in \emph{BMVC}, 2016.

\bibitem{Jain_cvpr14}
M.~Jain, J.~C. van Gemert, H.~J{\'{e}}gou, P.~Bouthemy, and C.~G.~M. Snoek,
  ``Action localization with tubelets from motion,'' in \emph{CVPR}, 2014.

\bibitem{Oneata_eccv14}
D.~Oneata, J.~Revaud, J.~J. Verbeek, and C.~Schmid, ``Spatio-temporal object
  detection proposals,'' in \emph{ECCV}, 2014.

\bibitem{FindingTube_GkioxariM15}
G.~Gkioxari and J.~Malik, ``Finding action tubes,'' in \emph{CVPR}, 2015.

\bibitem{YuYL13}
G.~Yu, J.~Yuan, and Z.~Liu, ``Action search by example using randomized visual
  vocabularies,'' \emph{{IEEE} Trans. Image Processing}, vol.~22, no.~1, pp.
  377--390, 2013.

\bibitem{FanSW13}
J.~Fan, X.~Shen, and Y.~Wu, ``What are we tracking: {A} unified approach of
  tracking and recognition,'' \emph{{IEEE} Trans. Image Processing}, vol.~22,
  no.~2, pp. 549--560, 2013.

\bibitem{ZhaoYHY15}
G.~Zhao, J.~Yuan, G.~Hua, and J.~Yang, ``Topical video object discovery from
  key frames by modeling word co-occurrence prior,'' \emph{{IEEE} Trans. Image
  Processing}, vol.~24, no.~12, pp. 5739--5752, 2015.

\bibitem{JiangMYL15}
Y.~Jiang, J.~Meng, J.~Yuan, and J.~Luo, ``Randomized spatial context for object
  search,'' \emph{{IEEE} Trans. Image Processing}, vol.~24, no.~6, pp.
  1748--1762, 2015.

\bibitem{JerripothulaCY16}
K.~R. Jerripothula, J.~Cai, and J.~Yuan, ``{CATS:} co-saliency activated
  tracklet selection for video co-localization,'' in \emph{ECCV}, 2016.

\bibitem{KangOLW16}
K.~Kang, W.~Ouyang, H.~Li, and X.~Wang, ``Object detection from video tubelets
  with convolutional neural networks,'' in \emph{CVPR}, 2016.

\bibitem{ActionProp_YuY15}
G.~Yu and J.~Yuan, ``Fast action proposals for human action detection and
  search,'' in \emph{CVPR}, 2015.

\bibitem{MaZIS13}
S.~Ma, J.~Zhang, N.~Ikizler{-}Cinbis, and S.~Sclaroff, ``Action recognition and
  localization by hierarchical space-time segments,'' in \emph{ICCV}, 2013.

\bibitem{Jan_bmvc15}
J.~C. van Gemert, M.~Jain, E.~Gati, and C.~G.~M. Snoek, ``{APT:} action
  localization proposals from dense trajectories,'' in \emph{BMVC}, 2015.

\bibitem{Li_accv16}
N.~Li, D.~Xu, Z.~Ying, and G.~L. Zhihao~Li, ``Search action proposals via
  spatial actionness estimation and temporal path inference and tracking,'' in
  \emph{ACCV}, 2016.

\bibitem{HochreiterS97}
S.~Hochreiter and J.~Schmidhuber, ``Long short-term memory,'' \emph{Neural
  Computation}, vol.~9, no.~8, pp. 1735--1780, 1997.

\bibitem{SelectiveSearch_UijlingsSGS13}
J.~R.~R. Uijlings, K.~E.~A. van~de Sande, T.~Gevers, and A.~W.~M. Smeulders,
  ``Selective search for object recognition,'' \emph{IJCV}, vol. 104, no.~2,
  pp. 154--171, 2013.

\bibitem{EdgeBox_ZitnickD14}
C.~L. Zitnick and P.~Doll{\'{a}}r, ``Edge boxes: Locating object proposals from
  edges,'' in \emph{ECCV}, 2014.

\bibitem{FasterRCNN_RenHGS15}
S.~Ren, K.~He, R.~B. Girshick, and J.~Sun, ``Faster {R-CNN:} towards real-time
  object detection with region proposal networks,'' in \emph{NIPS}, 2015.

\bibitem{YOLO_Redmon}
J.~Redmon, S.~Divvala, R.~Girshick, and A.~Farhadi, ``You only look once:
  Unified, real-time object detection,'' in \emph{CVPR}, 2016.

\bibitem{SSD_LiuAESRFB16}
W.~Liu, D.~Anguelov, D.~Erhan, C.~Szegedy, S.~E. Reed, C.~Fu, and A.~C. Berg,
  ``{SSD:} single shot multibox detector,'' in \emph{ECCV}, 2016.

\bibitem{KeSH05}
Y.~Ke, R.~Sukthankar, and M.~Hebert, ``Efficient visual event detection using
  volumetric features,'' in \emph{ICCV}, 2005.

\bibitem{LanWM11}
T.~Lan, Y.~Wang, and G.~Mori, ``Discriminative figure-centric models for joint
  action localization and recognition,'' in \emph{ICCV}, 2011.

\bibitem{TianSS13}
Y.~Tian, R.~Sukthankar, and M.~Shah, ``Spatiotemporal deformable part models
  for action detection,'' in \emph{CVPR}, 2013.

\bibitem{TranY12}
D.~Tran and J.~Yuan, ``Max-margin structured output regression for
  spatio-temporal action localization,'' in \emph{NIPS}, 2012.

\bibitem{PrimeProposal_ManenGG13}
S.~Manen, M.~Guillaumin, and L.~J.~V. Gool, ``Prime object proposals with
  randomized prim's algorithm,'' in \emph{ICCV}, 2013.

\bibitem{BroxM10_eccv10}
T.~Brox and J.~Malik, ``Object segmentation by long term analysis of point
  trajectories,'' in \emph{ECCV}, 2010.

\bibitem{RCNN_GirshickDDM14}
R.~B. Girshick, J.~Donahue, T.~Darrell, and J.~Malik, ``Rich feature
  hierarchies for accurate object detection and semantic segmentation,'' in
  \emph{CVPR}, 2014.

\bibitem{WeinzaepfelHS15}
P.~Weinzaepfel, Z.~Harchaoui, and C.~Schmid, ``Learning to track for
  spatio-temporal action localization,'' in \emph{ICCV}, 2015.

\bibitem{ILSVRC15}
O.~Russakovsky, J.~Deng, H.~Su, J.~Krause, S.~Satheesh, S.~Ma, Z.~Huang,
  A.~Karpathy, A.~Khosla, M.~Bernstein, A.~C. Berg, and L.~Fei-Fei, ``{ImageNet
  Large Scale Visual Recognition Challenge},'' \emph{International Journal of
  Computer Vision (IJCV)}, vol. 115, no.~3, pp. 211--252, 2015.

\bibitem{HeICCV15}
K.~He, X.~Zhang, S.~Ren, and J.~Sun, ``Delving deep into rectifiers: Surpassing
  human-level performance on imagenet classification,'' in \emph{ICCV}, 2015.

\bibitem{Adam}
\BIBentryALTinterwordspacing
D.~P. Kingma and J.~Ba, ``Adam: {A} method for stochastic optimization,''
  \emph{CoRR}, vol. abs/1412.6980, 2014. [Online]. Available:
  \url{http://arxiv.org/abs/1412.6980}
\BIBentrySTDinterwordspacing

\end{thebibliography}

\end{document}